\documentclass[letterpaper, 10 pt, conference]{ieeeconf}
\IEEEoverridecommandlockouts

\makeatletter
\let\labelindent\@undefined
\makeatother
\providecommand{\tablename}{TABLE}

\usepackage{cite}
\usepackage{amsmath,amssymb,amsfonts,amsthm}
\usepackage{graphicx}
\usepackage{textcomp}
\usepackage{xcolor}
\usepackage{bm}
\usepackage{booktabs,array,tabularx}
\usepackage{enumitem}
\usepackage{url}
\usepackage{dblfloatfix}
\usepackage[hidelinks,breaklinks=true]{hyperref}
\makeatletter

\def\@biblabel#1{[\raisebox{0pt}[0pt][0pt]{\hypertarget{Hcite.#1}{}}#1]}
\makeatother

\graphicspath{{figures/}}
\newcommand{\vect}[1]{\bm{#1}}
\newcommand{\R}{\mathbb{R}}
\newcommand{\SE}{\mathrm{SE}(3)}
\newcommand{\feas}{\mathrm{feas}}
\newcommand{\vis}{\mathrm{vis}}
\newtheoremstyle{vrw}{3pt plus 0pt minus 0pt}{3pt plus 0pt minus 0pt}{}{}{\bfseries}{.}{.5em}{}
\theoremstyle{vrw}

\AtBeginDocument{%
  \setlength{\abovedisplayskip}{4pt plus 2pt minus 2pt}%
  \setlength{\belowdisplayskip}{4pt plus 2pt minus 2pt}%
  \setlength{\abovedisplayshortskip}{2pt plus 1pt minus 1pt}%
  \setlength{\belowdisplayshortskip}{3pt plus 1pt minus 1pt}%
}
\newif\ifanonymize
\anonymizefalse

\newcommand{\ourrobot}{\ifanonymize our humanoid platform\else Duke Humanoid V2\fi}
\newcommand{\Ourrobot}{\ifanonymize Our humanoid platform\else Duke Humanoid V2\fi}
\newcommand{\mypara}[1]{\par\addvspace{1mm}\noindent\textbf{#1}}

\newif\ifinternalSupplement
\internalSupplementtrue  % Internal review build: render supplement.
\newcommand{\paperTitle}{Visible-Reachable Workspace for Perception-Aware Humanoid Design}
\newcommand{\tablecaption}[1]{%
  \refstepcounter{table}%
  \begingroup\footnotesize
  \leftskip=0pt \rightskip=0pt \parfillskip=0pt plus 1fil
  \noindent\MakeUppercase{\tablename}~\thetable: #1\par
  \endgroup
  \vspace{0.5\baselineskip}%
}

\title{\paperTitle}
\ifanonymize
  \author{Anonymous Authors}
\else
  \author{Boxi Xia, Zijiang Yang, Ryan Shin, Bokuan Li, Eric Wun-Hao Lu, Jacob Lee, Jiaxun Liu, Boyuan Chen%
  \thanks{*All authors are from Duke University. This work is supported by DARPA FoundSci program under award HR00112490372, DARPA TIAMAT program under award HR00112490419, ARO under award W911NF2410405, ARL STRONG program under awards W911NF2320182, W911NF2220113, and W911NF242021.}}
\fi

\makeatletter
\let\IEEEaftertitletext\@IEEESAVECMDIEEEaftertitletext
\makeatother

\IEEEaftertitletext{%
  \begin{center}
    \includegraphics[width=\linewidth]{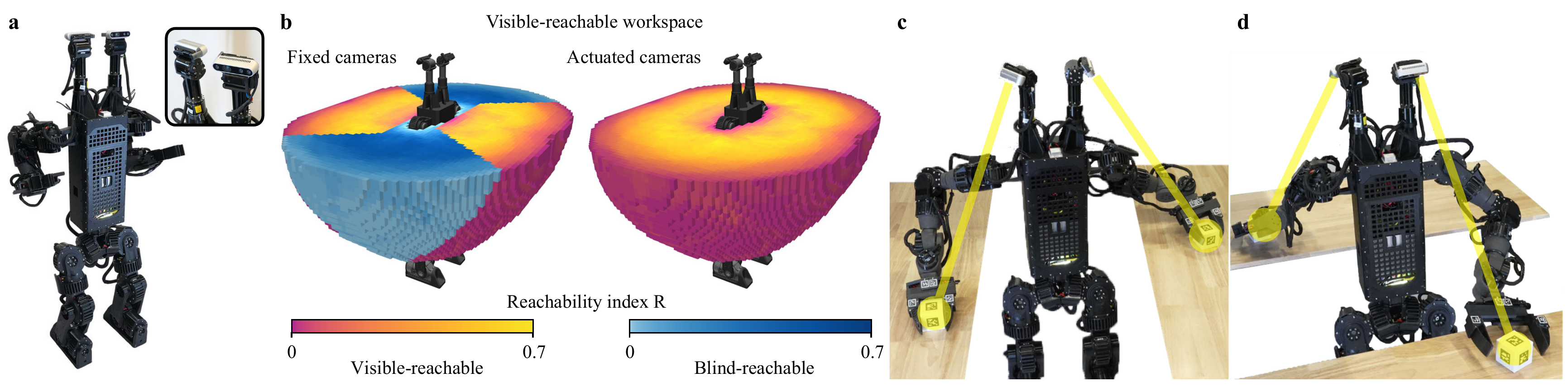}\\[4pt]
    \refstepcounter{figure}%
    \label{fig:teaser}%
    \parbox{\linewidth}{\footnotesize \textbf{Fig.~\thefigure.~From reachable space to visually usable manipulation space.} \textbf{(a)}~\Ourrobot{} carries two independently actuated yaw-pitch RGB-D cameras, allowing gaze to move independently of torso heading. \textbf{(b)}~Visible-reachable workspace (VRW) distinguishes arm-reachable voxels that are visible (plasma) from those that are reachable but blind (blue). Shade encodes orientation reachability $R$. On this platform, camera actuation raises visible-reachable coverage from $38\%$ to $97\%$. \textbf{(c,d)}~Two independent views can be assigned to spatially separated work regions during left/right and front/back manipulation.}
  \end{center}
}

\begin{document}

\pagenumbering{arabic}
\setcounter{page}{1}

% Standard IEEE conference paper starts here.
\maketitle

\begin{abstract}

Workspace analysis measures where a robot can place its end effector. For visually guided manipulation, reachability alone is insufficient: a kinematically reachable target may not be visible in the specific pose required to reach it. The robot must then redirect its sensing or move its body to acquire a view, turning a perception limitation into additional motion. Existing humanoids largely inherit this limitation when copying human form factors. We introduce the \emph{visible-reachable workspace} (VRW), a design-stage measure that conditions visibility on feasible reaching configurations and extends it to concurrent visibility of spatially separated work regions. We apply VRW by building a 31-DoF humanoid with independently actuated RGB-D cameras. On the same robot, camera articulation increases visible-reachable coverage from $38\%$ to $97\%$. With actuated camera layouts, a second camera raises pairwise coverage from $0.45$ to $0.95$, while a third changes it only to $0.97$. In a controlled two-target reach-and-grasp benchmark, our dual-actuated design reduces mean completion time by $17\%$ and mechanical energy by $19\%$ relative to the same robot with its cameras fixed. Hardware experiments demonstrate simultaneous observation and manipulation of front/back and left/right target pairs without torso reorientation. The results suggest that reachability becomes a more informative design quantity for perception-driven humanoid manipulation when it is evaluated together with the sensing configurations that make the reachable space observable. We will open-source all software and the humanoid hardware design.
Our website is \ifanonymize\url{https://anonymous.4open.science/w/visible-reachable-workspace/}\else\url{https://generalroboticslab.com/DukeHumanoidv2}\fi.
\end{abstract}

%==============================================================================
\section{Introduction}
\label{sec:introduction}

Reachability is one of the classical measures for describing a robot's manipulation capability~\cite{zacharias2007capturing,akinola2018workspace}. However, for a perception-driven humanoid robot that relies on onboard sensors, this geometric question is incomplete. A target can lie inside the arm workspace while remaining outside every available camera view at the configurations that realize the reach. Existing humanoids inherit this mismatch from human form factors, with broad arm workspaces paired with constrained vision concentrated in the head or the chest. The mechanism can therefore reach a target that the perception system cannot directly observe, leaving a perception-driven manipulation stack unable to localize, monitor, or plan around the target without first changing the sensing geometry.

This gap matters increasingly as humanoid control becomes more perception-dependent. Whether the downstream controller is model-based or learned, manipulation commonly relies on visual estimates of targets, obstacles, and task progress. When sensing and manipulation geometry are poorly matched, the robot must compensate by redirecting a camera, rotating its torso, or walking to a new viewpoint. Thus, a limitation that appears to be perceptual at design time can become whole-body motion at execution time.

The problem becomes more prominent in tasks spanning multiple work regions. For example, a robot may need to work across two benches or shelves, or interact with a person while monitoring another task. Steerable camera heads~\cite{saidi2007active,helpmate2000,icubvergence} can redirect gaze from one region to the other, but their cameras are aimed at a common target, providing only one viewing direction at a time. Therefore, one steerable view may provide high single-target visibility while still forcing sequential sensing or frequent body reorientation when two separated regions matter at once. Omnidirectional or distributed sensors provide broader passive coverage~\cite{omnimanip2026,armor2024,argus2026} but require more sensors and greater computational power.

We introduce the \emph{visible-reachable workspace} (VRW) to study sensing and manipulation capability as a coupled design principle. VRW starts from the conventional reachable workspace and retains only targets that can also be observed from a feasible reaching configuration. The definition is task-dependent, which means joints that are required to realize the active manipulation are distinguished from joints that remain independently available to aim cameras. This distinction is important for cameras placed on the manipulators. For example, a wrist-mounted camera could be included in the visibility model, but its pose is coupled to the arm configuration used for the current reach. Hence, it is not treated as an independently steerable gaze degree of freedom. In contrast, a camera gimbal that does not affect the end-effector pose can contribute independent sensing actuation. We further extend VRW to a manipulation target paired with a second region of interest, capturing whether a camera layout can maintain concurrent views of separated regions.

We use these measures to guide and evaluate the sensing layout of a 31-DoF quasi-direct-drive (QDD) humanoid with two independently actuated yaw-pitch RGB-D cameras (Fig.~\ref{fig:teaser}a). Across several existing humanoid embodiments, the analysis reveals a gap between what the arms can reach and what the onboard cameras can observe. Under our common geometric evaluation, the visible-reachable fraction ranges from $16\%$ for the fixed-head Unitree G1 to $48$--$76\%$ for platforms with an actuated neck, including PAL Talos, Booster T1, Fourier GR-3, and Apptronik Apollo. On \ourrobot{}, independently actuating the cameras raises this fraction from $38\%$ with the same cameras fixed to $97\%$ (Fig.~\ref{fig:teaser}\textbf{b}).

Single-target coverage, however, captures only part of the sensing problem. When two spatially separated regions must be observed, a steerable but shared viewing direction remains coupled. On \ourrobot, pairwise coverage rises from $0.45$ with one independently actuated camera to $0.95$ with two, while a third camera adds little, reaching $0.97$. The geometric deficit also appears during execution. In controlled two-target simulations, the dual-actuated configuration completes successful tasks $17\%$ faster and uses $19\%$ less mechanical energy than the same robot with its cameras fixed. The reduction occurs across search, approach, and manipulation time, suggesting a direct physical consequence of blind-but-reachable space. When the eyes cannot acquire the workspace, the body must move instead. Our physical experiments reproduce the intended dual-region behavior in front/back and left/right manipulation (Fig.~\ref{fig:teaser}\textbf{c,d}).

%--- reachable-and-visible workspace: reach cap visible -------------
\begin{figure*}[!t]
\centering
\includegraphics[width=\linewidth]{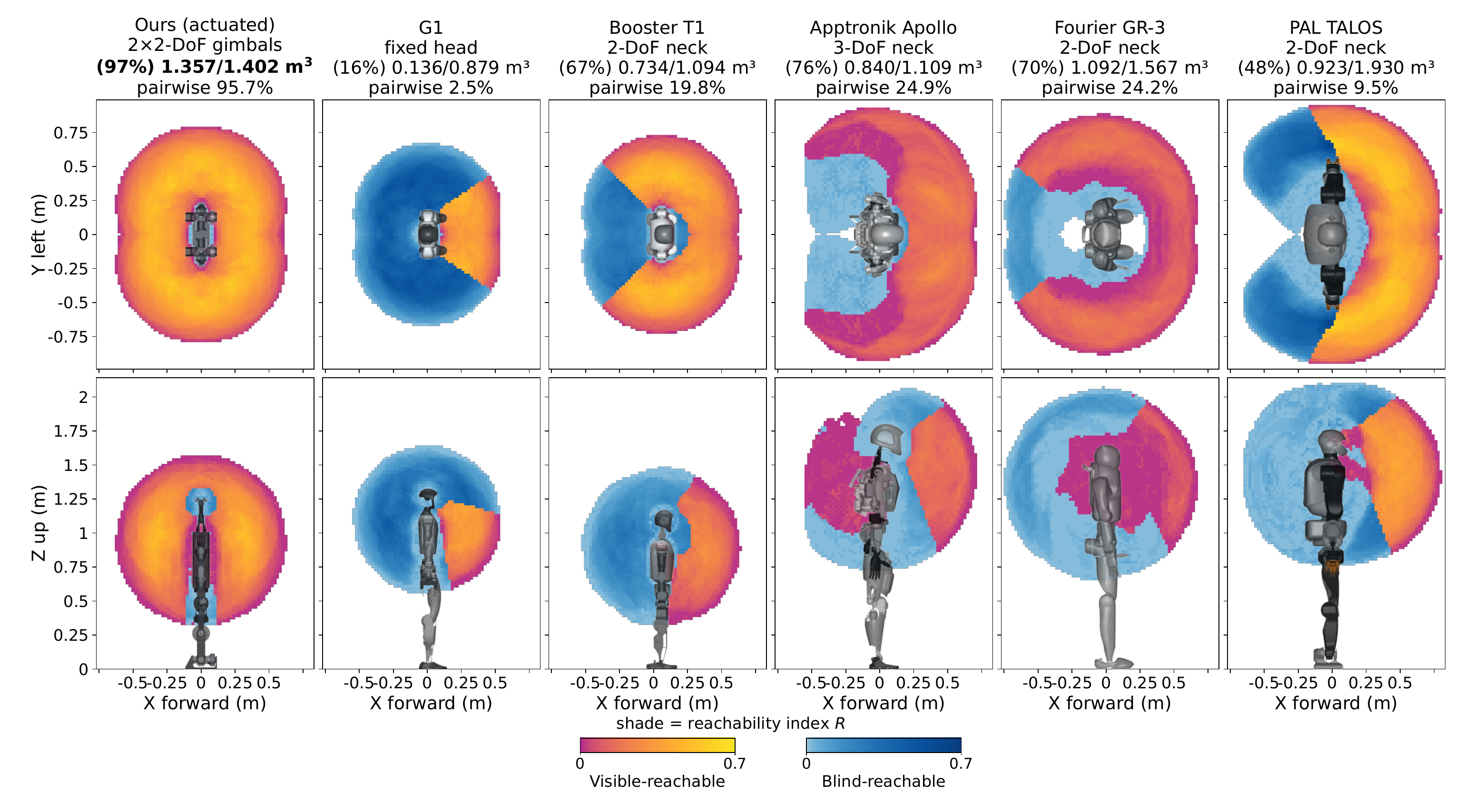}
\vspace{-24pt}
\caption{\textbf{Visible-reachable workspace.} Collision-free IK is solved at each $20$~mm target grid point for $64$ near-uniform SO(3) orientations~\cite{alexa2022super}. Color encodes the orientation reachability index $R$, the fraction of those orientations that admit a solution. Columns compare our actuated configuration (Act$_2$ of Section~\ref{sec:results}), Unitree G1, Booster T1, Apptronik Apollo, Fourier GR-3, and PAL Talos, with the external models taken from mjlab~\cite{zakka2026mjlab}, Fourier Intelligence's GRx repository~\cite{fourier_grx_models}, and MuJoCo Menagerie~\cite{mujoco_menagerie}. Top: $X$--$Y$ section at shoulder height. Bottom: $X$--$Z$ cutaway $y\geq0$, each cell showing the frontmost voxel nearest the cut plane. Plasma and blue denote visible-reachable and blind voxels. Each heading reports visible percentage, visible-reachable and reachable volumes in m$^3$, camera articulation, and pairwise coverage $\eta_2$. $\tilde\eta$ is the VRW coverage computed with the canonical arms-down occlusion model. None of the external platforms can aim two camera views independently.}
\label{fig:workspace}
\end{figure*}

%==============================================================================
\section{Related Work}
\label{sec:related}

\mypara{Reachability as a robot capability.}
Workspace and capability maps summarize where a manipulator can reach and, in richer formulations, which end-effector orientations are feasible~\cite{zacharias2007capturing,akinola2018workspace}. These representations are useful for mechanism evaluation, base placement, and grasp planning, but the usual reachable set does not encode whether an onboard sensor can observe the target from a feasible reaching configuration. For visually guided manipulation, this omission can make geometric reachability an optimistic measure of locally usable workspace.

\mypara{Visibility-aware manipulation and active vision.}
Prior work has coupled visibility with action at planning or control time. Mightability maps combine reachability and visibility for human-robot perspective taking~\cite{pandey2010mightability}. MoMaGen checks reachability and visibility while sampling base and head configurations~\cite{li2026momagen}, and visibility-aware controllers explicitly preserve line of sight during whole-body motion~\cite{he2022visibility}. Active-vision systems redirect a camera, stereo pair, or pan-tilt head toward task-relevant regions~\cite{saidi2007active,eyerobot2025,halomi2026,avaloha2024,activestereo2026}, including convergent gaze systems such as HelpMate~\cite{helpmate2000} and iCub~\cite{icubvergence}. These methods demonstrate the value of moving the sensor during a task. Our emphasis is complementary. We evaluate the sensor arrangement itself, before execution, by asking how its coverage range overlaps the robot's manipulation capability.

\mypara{Sensor layout and humanoid perception.}
Humanoid platforms span human-mimetic, compliant, learning-oriented, and locomotion-efficient designs~\cite{asano2017design,valkyrie,bostondynamics2026atlas,toddlerbot2025,berkeleylite,berkeleyhumanoid,xia2025duke}. Their onboard perception is commonly fixed to the torso or head, or moved through a shared neck. Other systems broaden coverage with omnidirectional or distributed sensing~\cite{omnimanip2026,armor2024}, while Argus uses active multi-camera perception on a non-humanoid platform~\cite{argus2026}. Camera-planning and workcell-design methods optimize camera poses or fixed coverage using scene and optical constraints~\cite{tarabanis1995mvp,oscadal2023camera}. We propose a design-stage measure of how humanoid onboard camera placement and articulation interact with the workspace of its arms, including whether separated regions can be viewed concurrently.

%==============================================================================
\section{Visible-Reachable Workspace}
\label{sec:vrw}

VRW begins with the classical fixed-base reachable workspace and asks which of those reaches are supported by the robot's onboard sensing. The floating base is fixed at a nominal stance during workspace analysis. This excludes locomotion and isolates the relationship between the robot's arm reach and onboard sensing from a given body pose. Allowing the base to move would confound this geometric property with the robot's ability to acquire a new viewpoint by walking, which we evaluate separately in the dynamic experiments (Section~\ref{sec:results}).

\mypara{Task-relative configuration partition.} Let $q\in\mathcal{Q}\subset\R^n$ denote the robot configuration and let $\mathcal{Q}^{\feas}$ contain configurations satisfying the joint limits and self-collision constraints used by the inverse-kinematics solver. Consider a task with active end-effectors $A$ (one hand for unimanual analysis and both hands when evaluating the union of their local workspaces). For end effector $i$, let $f_i(q)\in\R^3$ denote its position, and for camera $k$ let $g_k(q)\in\SE$ denote its pose.

The degrees of freedom available for sensing depend on the manipulation being evaluated. We therefore partition the configuration as $ q=(q_m,q_p,q_0)$, where $q_m$ contains every joint on which any active end-effector pose depends, $q_p$ contains the remaining joints on which at least one camera pose depends, and $q_0$ contains all other joints, held at specified nominal values. The assignment gives manipulation priority. If a joint affects both an active end effector and a camera, it belongs to $q_m$ rather than $q_p$.

This partition clarifies the treatment of manipulation-coupled sensing. A wrist-mounted camera is not excluded from VRW. Its visibility can be evaluated through $g_k(q_m)$, but because the camera moves with the active arm, it contributes no independent $q_p$ actuation. Its gaze can change only through alternative feasible reaching configurations. Conversely, a pan-tilt unit whose joints do not affect the end effector belongs to $q_p$ and can be aimed after the manipulation configuration is fixed. Our experiments focus on head-mounted RGB-D cameras because the design question is independent gaze for separated work regions, but the formulation itself does not require that mounting choice. Throughout this section, $(q_m,q_p)\in\mathcal{Q}^{\feas}$ abbreviates the full configuration $(q_m,q_p,q_0)$ with $q_0$ fixed to its nominal value.

\mypara{From reachable to visible-reachable workspace.} For active end effector $i\in A$, the conventional reachable workspace is
\begin{equation}
\mathcal{W}^{(i)}_R=
\left\{x\in\R^3\;\middle|\;\exists(q_m,q_p)\in\mathcal{Q}^{\feas}: f_i(q_m)=x\right\}.
\label{eq:reachable-workspace}
\end{equation}
Because $f_i$ does not depend on $q_p$, Eq.~\eqref{eq:reachable-workspace} is the usual manipulation workspace over the task-relevant kinematic chain, subject to the complete robot's feasibility constraints. Camera modules remain in the collision model, so changing the sensor layout can slightly change $\mathcal{W}^{(i)}_R$. As in capability-map analysis, we additionally report an orientation reachability index $R(x)$, the fraction of sampled end-effector orientations at $x$ for which collision-free IK succeeds.

For a specific reaching configuration $q_m$, define conditional visibility by
\begin{equation}
\begin{split}
V(x;q_m)=1 \iff &\exists q_p,k:\ (q_m,q_p)\in\mathcal{Q}^{\feas},\\
&\vis(x;k,q)=1,
\end{split}
\label{eq:conditional-visibility}
\end{equation}
where $\vis(x;k,q)=1$ when $x$ lies within camera $k$'s modeled field of view and range and the camera-to-target segment is not self-occluded. This conditioning is the key distinction from an unconstrained visibility map. The robot receives credit only for a view compatible with a configuration that realizes the manipulation. Independent sensing joints may move through $q_p$; manipulation-coupled sensor motion is inherited from $q_m$ and cannot be optimized as a separate gaze action.

The visible-reachable workspace is then
\begin{equation}
\mathcal{W}^{(i)}_{VR}=\left\{x\in\R^3\;\middle|\;\exists q_m:\ f_i(q_m)=x,\ V(x;q_m)=1\right\},
\label{eq:vrw}
\end{equation}
with coverage
\begin{equation}
\eta=\frac{\mu(\mathcal{W}^{(i)}_{VR})}{\mu(\mathcal{W}^{(i)}_R)}\in[0,1].
\label{eq:eta}
\end{equation}
Here $\mu$ denotes volume, estimated by occupancy on a discretized workspace grid. The complement $\mathcal{W}^{(i)}_R\setminus\mathcal{W}^{(i)}_{VR}$ is blind-reachable: kinematically reachable, but not observable by onboard sensing from any compatible configuration under the stated task partition.

For dense cross-layout visualization, we use a canonical-occlusion approximation $\tilde\eta$. Self-occlusion is evaluated with the arms in a common arms-down pose while camera articulation and all other sensor-model constraints are retained. This approximation makes the maps consistent and computationally tractable, but does not retain the reach-conditioned occlusion in Eq.~\eqref{eq:conditional-visibility}. Unless otherwise noted, the reported VRW coverage values use $\tilde\eta$.

For whole-robot maps we report the union $\mathcal{W}_R^A=\bigcup_{i\in A}\mathcal{W}^{(i)}_R$ and $\mathcal{W}_{VR}^A=\bigcup_{i\in A}\mathcal{W}^{(i)}_{VR}$, with $A$ containing the two hands. This union summarizes local coverage of either hand and should not be interpreted as simultaneous bimanual feasibility.

\mypara{Concurrent visibility of separated regions.} Single-target VRW can make one- and two-camera systems appear nearly equivalent when either system can eventually direct a view toward an individual target. This equivalence breaks down in multi-region manipulation, where the relevant capability is not eventual visibility of each target in isolation, but \textbf{concurrent visibility:} maintaining observation of the active manipulation region while simultaneously monitoring a second region without altering the manipulation configuration.

For a reachable manipulation target $x_i$ and a second region-of-interest point $x_j$, define
\begin{equation}
\begin{split}
C_i(x_i,x_j)=1 \iff
&\exists(q_m,q_p)\in\mathcal{Q}^{\feas},\ \exists k_i,k_j:\\
&f_i(q_m)=x_i,\quad \vis(x_i;k_i,q)=1,\\
&\vis(x_j;k_j,q)=1.
\end{split}
\label{eq:concurrent}
\end{equation}
The cameras $k_i$ and $k_j$ need not be distinct. A single camera therefore receives credit when both regions fit inside one unoccluded frustum, whereas independently actuated cameras can instead direct different views toward the two regions.

We summarize this property by
\begin{equation}
\eta_2=\Pr_{(x_i,x_j)\sim\nu_i\times\nu_j}\!\left[C_i(x_i,x_j)=1\right].
\label{eq:eta2}
\end{equation}
The distribution $\nu_i$ samples reachable grid points in proportion to orientation reachability $R(x_i)$, emphasizing locations that admit more grasp orientations, while $\nu_j$ samples the specified evaluation volume uniformly. The second point is intentionally not required to be reachable. $\eta_2$ asks whether the robot can manipulate in one region while maintaining perception of another. If a task instead requires simultaneous bimanual manipulation, the same construction can restrict $x_j$ to the second arm's reachable set and impose its reach constraint explicitly.

% For comparison with single-target visibility, we also report $\eta_1=\Pr_{x_i\sim\nu_i}[x_i\in\mathcal{W}^{(i)}_{VR}]$ and retention $\rho=\eta_2/\eta_1$. 
The scalar $\eta_2$ is distribution dependent and should not be interpreted as a universal cross-platform ranking. Different embodiments induce different evaluation volumes and target-separation distributions. Figure~\ref{fig:workspace} therefore uses cross-platform values only as descriptive context. The camera-count study in Section~\ref{sec:results:workspace} provides the controlled comparison, evaluating all layouts on the same robot using a common grid and separation bins.

%==============================================================================
\section{Humanoid Realization}
\label{sec:platform}

The VRW metric is embodiment-agnostic. We evaluate it on various platforms and present Duke Humanoid V2 for systematic design studies that demonstrate the effectiveness of this sensing-manipulation design metric.

\begin{figure}[t]
\centering
\includegraphics[width=0.95\linewidth]{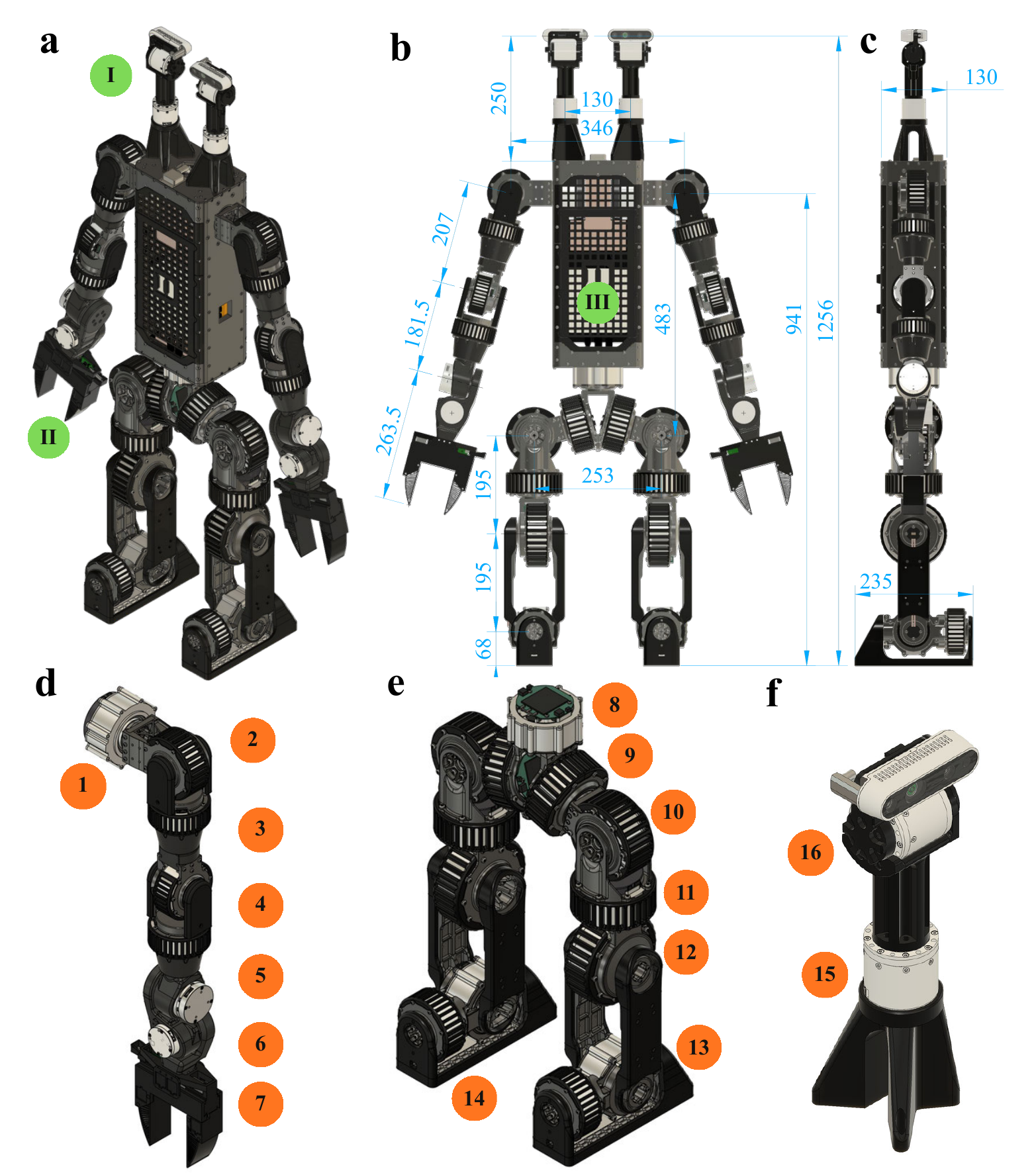}
\caption{\textbf{\Ourrobot{} with individually controlled dual cameras.} Orange numbers indicate actuated joints: (1)~shoulder pitch, (2)~shoulder roll, (3)~shoulder yaw, (4)~elbow, (5)~wrist roll, (6)~wrist pitch, (7)~wrist yaw, (8)~waist, (9)~hip pitch, (10)~hip roll, (11)~hip yaw, (12)~knee, (13)~ankle pitch, (14)~ankle roll, (15)~camera yaw, (16)~camera pitch. Green labels identify hardware modules: (I)~camera module, (II)~gripper, (III)~onboard computer. \textbf{(a)}~27-DoF body with 4-DoF dual-camera head. \textbf{(b)}~Front view (mm). \textbf{(c)}~Side view (mm). \textbf{(d)}~7-DoF arm and parallel gripper. \textbf{(e)}~Lower body. \textbf{(f)}~Yaw-pitch camera gimbal.}
% \vspace*{-10pt}
\label{fig:platform}
\end{figure}

\Ourrobot{} is a 31-DoF QDD humanoid comprising one waist joint, two 6-DoF legs, and two 7-DoF arms, plus two 2-DoF camera gimbals (Fig.~\ref{fig:platform}). Each parallel gripper adds another actuated DoF. The robot is 1.2~m tall, has a mass of 36~kg, a 0.46~m arm reach, and a leg length of 0.39~m. The robot uses CNC-machined 6061 aluminum and 3D-printed Nylon~12 for the main structure, PLA for electronics mounts, and TPU for protection.

% \subsection{Control-policy architecture}
\begin{figure*}[!t]
\centering
\includegraphics[width=0.95\linewidth]{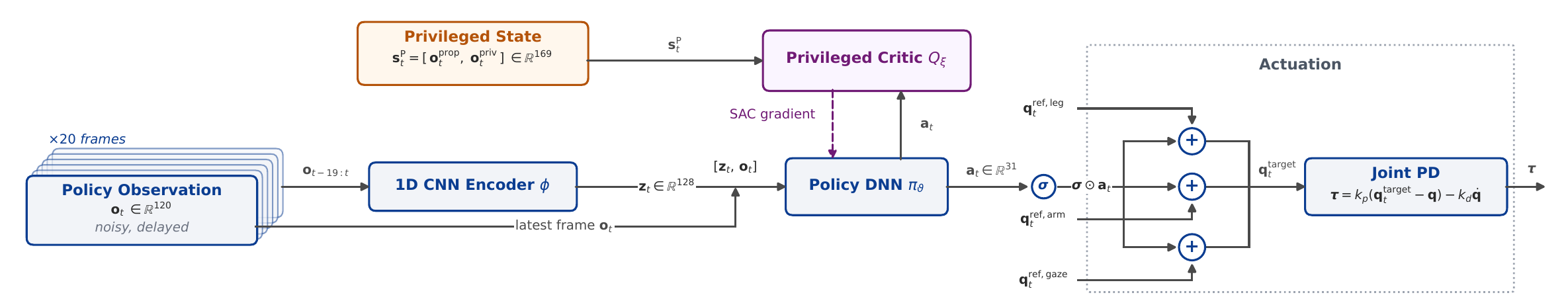}
\vspace*{-4mm}
\caption{\textbf{Control architecture.} We train the whole-body policy using FlashSAC~\cite{flashsac2026}. The actor encodes a $20$-frame observation history and outputs residual joint-position commands for the legs, arms, and cameras. The residual commands are added to the reference joint positions and tracked by joint-level PD control. The asymmetric critic (dashed) uses privileged information only during training.}
\label{fig:control_architecture}
% \vspace*{2mm}
\end{figure*}

% The actor embeds each $120$-D observation frame ($120\!\to\!32$), applies three 1D convolutions (channels $32\!\to\!32\!\to\!128$, kernel $3$), mean-pools the $20$-frame history into $\vect{z}\in\mathbb{R}^{128}$, and feeds $\vect{z}$ with the current frame to an MLP trunk ($512\!\to\!256\!\to\!128$) and a tanh-squashed Gaussian head. A twin distributional C51 asymmetric critic uses two $101$-atom heads over $[-5,5]$ with clipped double-Q.

\begin{table}[t]
    \centering
    \footnotesize
    \tablecaption{\Ourrobot{} per-joint ROM and max torque (N$\cdot$m). Values are for the left-side joints, mirrored on the right.}
    \label{tab:joint_spec}
    \setlength{\tabcolsep}{2pt}
    \resizebox{\linewidth}{!}{%
    \begin{tabular}{l c c @{\hspace{2pt}\vrule width 0.8pt\hspace{2pt}} l c c}
        \hline
        \textbf{Joint} & \textbf{ROM ($^\circ$)} & \textbf{Torque} & \textbf{Joint} & \textbf{ROM ($^\circ$)} & \textbf{Torque} \\
        \hline
        Waist yaw      & $[-90,90]$    & $60$  & Shoulder roll & $[-180,30]$   & $36$  \\
        Hip pitch      & $[-105,105]$  & $60$  & Shoulder yaw  & $[-180,180]$  & $17$  \\
        Hip roll       & $[-105,30]$   & $60$  & Elbow         & $[-125,125]$  & $17$  \\
        Hip yaw        & $[-90,90]$    & $60$  & Wrist roll    & $[-180,180]$  & $17$  \\
        Knee           & $[-130,130]$  & $120$ & Wrist pitch   & $[-92,92]$    & $14$  \\
        Ankle pitch    & $[-50,50]$    & $60$  & Wrist yaw     & $[-90,90]$    & $5.5$ \\
        Ankle roll     & $[-60,60]$    & $36$  & Camera yaw    & $[-270,270]$  & $5.5$ \\
        Shoulder pitch & $[-108,108]$  & $60$  & Camera pitch  & $[-90,90]$    & $5.5$ \\
        \hline
    \end{tabular}%
    }
\vspace*{-8pt}
\end{table}

\mypara{Mechanical layout.} The shoulder pitch motor is packaged inside the torso to reduce shoulder width, and the shoulder/elbow ranges permit reaches in front of and behind the torso. The lower body uses an offset hip-roll axis and approximately symmetric sagittal ranges to support forward, lateral, and backward locomotion. This symmetric design for both locomotion and manipulation supports more versatile whole-body motion with less need for body reorientation.

\mypara{Actuation and end effectors.} All non-gripper joints use QDD motors (RobStride, 7.75:1--10:1 reduction, dual 14-bit encoders). Per-joint range of motion (ROM) and motor torque are listed in Table~\ref{tab:joint_spec}. Locomotion uses an IMU (TransducerM TM171) and motor state feedback. An onboard MINISFORUM X1-470 computer communicates with the actuators over CAN. Each wrist carries a parallel gripper (350~g, 184~mm maximum opening) that uses UMI-inspired finger geometry~\cite{chi2024universal}. We mount eight AprilTags on each gripper for pose tracking. Workspace IK is referenced to the grasp center.

\mypara{Independently actuated sensing.}
Two yaw-pitch camera modules are mounted on top of the robot. Each module uses an Intel RealSense D436 RGB-D camera with a $90^\circ\!\times65^\circ$ RGB field of view and a $0.1$--$3.0$~m sensing range. Each module is actuated independently, allowing the two cameras to simultaneously view spatially separated regions. Section~\ref{sec:results:workspace} evaluates alternative camera configurations and whether adding an independent view justifies the additional hardware cost.

%==============================================================================
\mypara{Control policy.} A whole-body RL policy runs at $50$~Hz with $200$~Hz joint-level PD tracking. We train it with FlashSAC~\cite{flashsac2026}, an off-policy soft actor-critic method~\cite{sac2018} with an asymmetric critic. A 1D CNN actor encodes a 20-frame history of noisy, delayed proprioception together with base-twist, arm, and gaze references (Fig.~\ref{fig:control_architecture}). Each frame is embedded from 120 to 32 dimensions and then processed by three 1D convolutions and mean pooling over the history. The resulting 128-D latent and the current observation are passed to an MLP that outputs the residual joint-position action. The critic additionally receives privileged simulator quantities during training. The 31-D residual joint-position action is added to group-specific references for the legs, planned arm waypoints, and left/right yaw-pitch gaze targets.

The command curriculum spans $v_x,v_y\!\in\![-1,1]$~m/s and $\omega_z\!\in\![-1,1]$~rad/s. The reward is a weighted sum of task-tracking terms (base linear and angular velocity, arm pose, and gaze), posture and balance, gait shaping (swing timing, clearance, slip, and impact), and regularization (action smoothness, joint limits, self-collision, and energy). Training uses $4096$ parallel environments for $15$k iterations in mjlab~\cite{zakka2026mjlab}, randomizing contact parameters, PD gains, body and wrist-payload inertia, and per-DoF friction, with periodic pushes.

%==============================================================================
\section{Results}
\label{sec:results}

\Ourrobot{} has a symmetric reachable workspace. To provide the camera coverage needed for visual manipulation, we used VRW to guide the design of the camera configuration. We considered two questions: \textbf{Q1}: How can VRW guide the design of the camera configuration? \textbf{Q2}: Does designing for higher VRW improve visual grasping performance?

\subsection{Camera configuration optimization}
\label{sec:results:workspace}

\begin{figure}[!t]
\centering
\vspace{6pt}
\includegraphics[width=\linewidth]{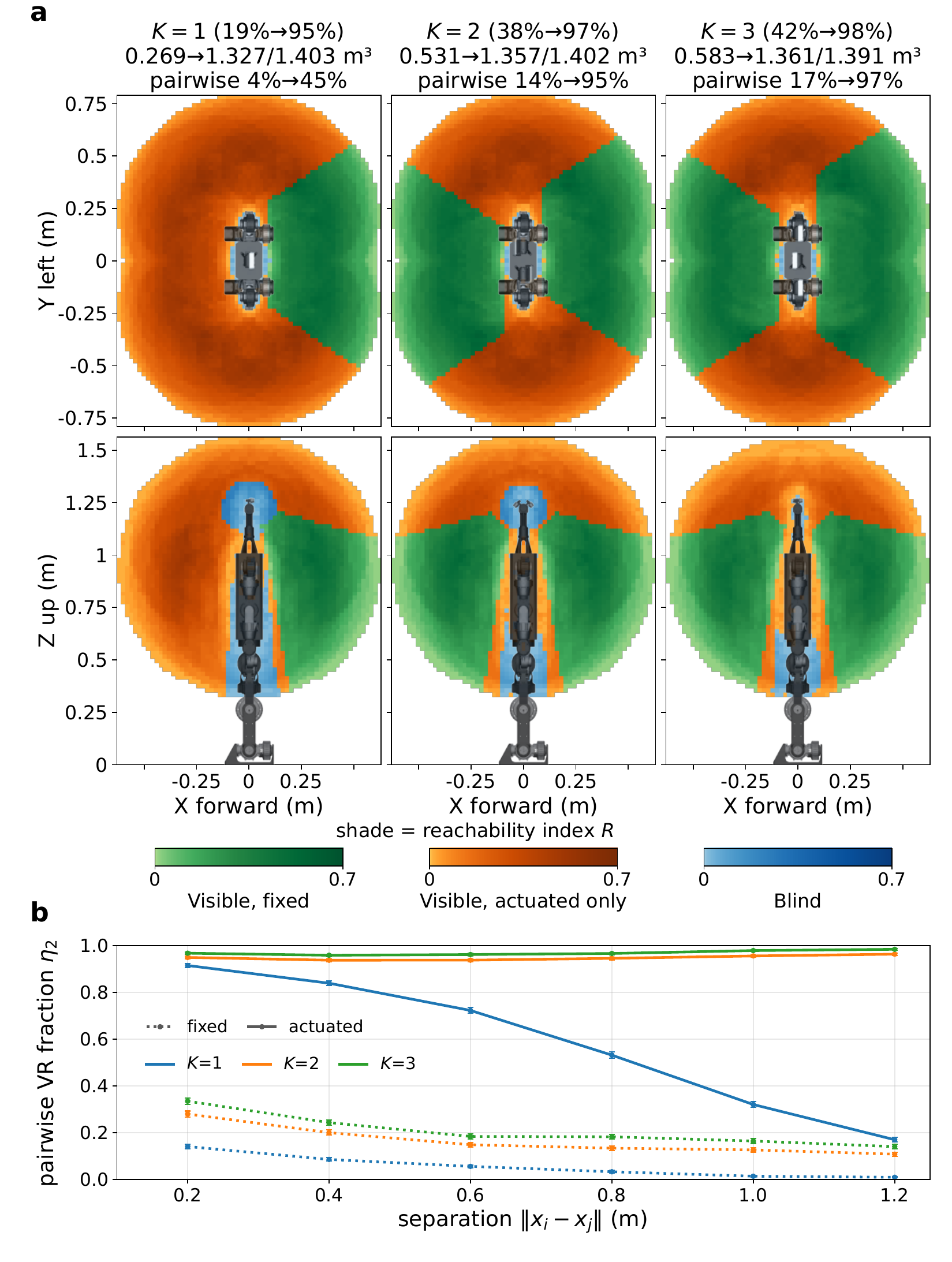}
\vspace{-20pt}
\caption{\textbf{Camera configurations.} \textbf{(a)}~Visible-reachable workspace for the $K=1$, $2$, and $3$ camera configurations. Top: $X$--$Y$ section at shoulder height. Bottom: $X$--$Z$ cutaway for $y\geq0$. Green denotes voxels visible with fixed camera joints, orange voxels visible only by actuating the camera joints, and blue blind voxels. Shade encodes reachability index $R$. Headings show fixed-to-actuated coverage, $\tilde\eta$, and scalar pairwise coverage. $K=2$ and $K=3$ appear similar because the third camera module adds lateral voxels outside these cut planes. \textbf{(b)}~Pairwise coverage $\eta_2$ against target separation, with $95\%$ Wilson intervals. Solid lines and dotted curves indicate actuated and fixed camera configurations, respectively.}
\label{fig:camera-count}
\vspace{-6pt}
\end{figure}

On \ourrobot{}, cameras could be mounted on the front, back, or top faces of the body. The front and back faces were not chosen because their visual coverage was limited to one side of the humanoid, whereas top-mounted cameras could cover the front, back, left, and right sides. We also did not consider wrist cameras because a wrist camera on the reaching arm offered no independent $q_p$ actuation under the partition of Section~\ref{sec:vrw}.

The remaining design decisions were camera count and articulation. We evaluated six layouts (Fig.~\ref{fig:camera-count}\textbf{a}): $K=1$, $2$, and $3$ camera modules, each either actuated (Act$_K$) or fixed (Fix$_K$). For the fixed layouts, $K=2$ used one front-facing and one back-facing camera, while $K=3$ used two front-facing cameras and one back-facing camera. All other robot configurations remained the same. We compared both single-target visible-reachable coverage $\tilde\eta$ and pairwise coverage $\eta_2$.
% We also compared our design with existing humanoid designs using the same VRW metric.

\mypara{Single-target coverage.} Camera articulation had a larger effect on VRW coverage than camera count (Fig.~\ref{fig:camera-count}\textbf{a}). Across all camera counts, actuating the cameras greatly increased VRW coverage. Once actuated, VRW coverage was already close to saturation with one camera module, and adding a second or third camera provided $<4\%$ coverage gains. 

\mypara{Pairwise coverage.} VRW coverage considers one manipulation target at a time and does not capture whether two targets can be viewed simultaneously. Camera count had a larger effect on pairwise coverage (Fig.~\ref{fig:camera-count}\textbf{b}). With one camera, both regions must lie within the same unoccluded camera view, so pairwise coverage decreased as target separation increased. With two independently actuated cameras, each camera could maintain a separate view, and pairwise coverage remained nearly constant as target separation increased. Actuating the cameras increased VRW coverage more than adding more cameras, while adding a second independently actuated camera increased pairwise coverage. Adding a third camera provided only a small further increase in pairwise coverage and visible-reachable volume, while adding $0.58$~kg, two gimbal DoF, and \$600 in hardware cost. Thus, we adopted the $K=2$ configuration with independently actuated camera modules.

\mypara{Comparison with other humanoids.} We also evaluated VRW on the fixed-head Unitree G1 and humanoids with actuated neck-mounted cameras (Booster T1, Apptronik Apollo, Fourier GR-3, and PAL Talos) (Fig.~\ref{fig:workspace}). We used the published camera field of view when available. For G1, we used our $90^\circ{\times}65^\circ$ camera model. For Apollo, we used a provisional $91.5^\circ{\times}60^\circ$ model. The fixed-head G1 had much lower VRW coverage, while the platforms with actuated necks covered a larger portion of their reachable workspace, although all remained below our design. 
For pairwise coverage, the external platforms remained below our design, including those with actuated necks. Their cameras share the same neck joints and cannot aim independently at two separated regions. GR-3 had the widest camera field of view among these platforms, but still provided only one viewing direction and therefore had lower pairwise coverage than our design.

\begin{figure}[!t]
\centering
\includegraphics[width=\linewidth]{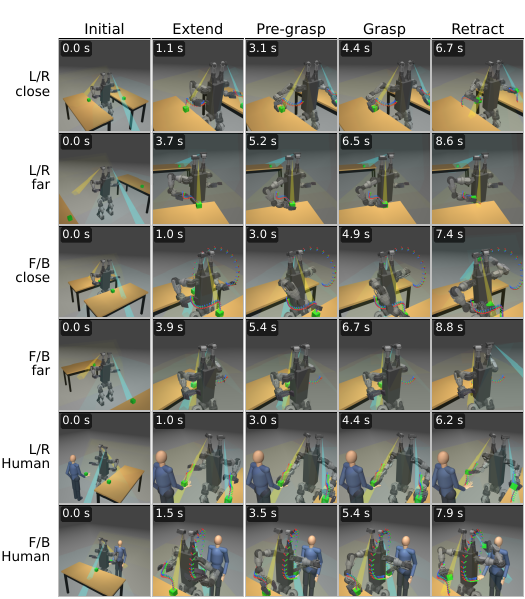}
\vspace{-12pt}
\caption{\textbf{Time-lapse frames of \ourrobot{} performing the two-target reach-and-grasp benchmark.} Rows: the six scenarios of Table~\ref{tab:scenarios-dynamic}. Columns: reach-and-grasp keyframes for one target: initial stage, arm extension, pre-grasp, grasp, and retraction. The targets are $60$\,mm cubes shown in green. Shaded cones show the camera fields of view, cyan for the left module and orange for the right. The narrow, more transparent cone shows the central $10\%$ of the field of view about the optical axis; the wider cone shows the full field of view, truncated at the camera near plane. Dotted coordinate frames show the planned collision-free arm trajectories.}
\label{fig:benchmark}
\vspace{-6pt}
\end{figure}

%------------------------------------------------------------------------------
%  Dynamic (full-physics) two-target evaluation
%------------------------------------------------------------------------------
\begin{table*}[!t]
    \centering
    \vspace*{5pt}
    \scriptsize
    \tablecaption{Two-target reach-and-grasp in physics  simulation. Each entry reports the average of $30$ trials ($10$ target layouts $\times$ $3$ trials). Fix$_K$/Act$_K$ denote fixed/actuated $K$-camera configurations, and G1 is the external baseline. \emph{Human} indicates one target held by a person. Time and energy are averaged over successful trials. Bold marks the best value for each metric.}
    \label{tab:scenarios-dynamic}
    \setlength{\tabcolsep}{1.4pt}   % 25 numeric columns: at the default 6pt/side, padding alone spends ~310pt
    \begin{tabular}{@{}l @{\hspace{0.3em}} ccccc @{\hspace{0.8em}} ccccc @{\hspace{0.8em}} ccccc @{\hspace{0.8em}} ccccc @{\hspace{0.8em}} ccccc@{}}
        \toprule
        & \multicolumn{5}{c}{\textbf{Time} $\bar{T}$ \textbf{(s) $\downarrow$}} & \multicolumn{5}{c}{\textbf{Search time} $\bar{T}_{\text{search}}$ \textbf{(s)} $\downarrow$} & \multicolumn{5}{c}{\textbf{Approach time} $\bar{T}_{\text{approach}}$ \textbf{(s)} $\downarrow$} & \multicolumn{5}{c}{\textbf{Manipulation time} $\bar{T}_{\text{manip}}$ \textbf{(s)}$\downarrow$} & \multicolumn{5}{c}{\textbf{Energy} $E$ \textbf{(J)}$\downarrow$} \\
        \cmidrule(lr){2-6}\cmidrule(lr){7-11}\cmidrule(lr){12-16}\cmidrule(lr){17-21}\cmidrule(lr){22-26}
        \textbf{Scenario} & G1 & Fix$_2$ & \textbf{Act$_2$} & Fix$_1$ & Act$_1$ & G1 & Fix$_2$ & \textbf{Act$_2$} & Fix$_1$ & Act$_1$ & G1 & Fix$_2$ & \textbf{Act$_2$} & Fix$_1$ & Act$_1$ & G1 & Fix$_2$ & \textbf{Act$_2$} & Fix$_1$ & Act$_1$ & G1 & Fix$_2$ & \textbf{Act$_2$} & Fix$_1$ & Act$_1$ \\
        \midrule
        L/R close & 29.3 & 15.9 & \textbf{10.6} & 20.7 & 12.8 & 3.2 & 0.3 & \textbf{0.1} & 2.8 & 0.3 & 5.9 & 1.9 & 0.2 & 3.5 & \textbf{0.0} & 20.2 & 13.7 & \textbf{10.3} & 14.4 & 12.5 & 575 & 264 & 81 & 514 & \textbf{36} \\
        L/R far & 34.2 & 25.8 & \textbf{24.2} & 28.8 & 26.0 & 7.2 & 0.7 & \textbf{0.3} & 6.3 & 1.2 & \textbf{6.4} & 8.2 & 6.6 & 7.0 & 8.2 & 20.6 & 16.9 & 17.3 & \textbf{15.5} & 16.6 & \textbf{821} & 1059 & 978 & 1201 & 1095 \\
        \addlinespace
        F/B close & 24.8 & 10.7 & 10.6 & 18.5 & \textbf{10.5} & 2.9 & 0.1 & \textbf{0.0} & 3.3 & 0.6 & 1.1 & \textbf{0.0} & \textbf{0.0} & 0.9 & \textbf{0.0} & 20.8 & 10.6 & 10.6 & 14.3 & \textbf{9.9} & 315 & 67 & 70 & 403 & \textbf{59} \\
        F/B far & 31.7 & 23.6 & \textbf{22.4} & 26.3 & 25.9 & 5.1 & \textbf{0.0} & 0.1 & 3.8 & 0.7 & 6.3 & 6.9 & \textbf{5.8} & 6.7 & 8.6 & 20.3 & 16.7 & 16.5 & \textbf{15.8} & 16.6 & \textbf{671} & 825 & 880 & 989 & 1078 \\
        \addlinespace
        L/R Human & 19.9 & 15.2 & \textbf{7.7} & 9.6 & 8.3 & 1.2 & 0.4 & 0.2 & \textbf{0.1} & \textbf{0.1} & 1.8 & 1.7 & \textbf{0.0} & 0.4 & \textbf{0.0} & 16.9 & 13.1 & \textbf{7.5} & 9.1 & 8.2 & 250 & 236 & 24 & 50 & \textbf{22} \\
        \addlinespace
        F/B Human & 24.8 & 10.7 & \textbf{9.6} & 18.8 & 10.2 & 2.9 & 0.1 & \textbf{0.0} & 3.3 & 0.6 & 1.3 & \textbf{0.0} & \textbf{0.0} & 0.8 & \textbf{0.0} & 20.6 & 10.6 & \textbf{9.6} & 14.7 & \textbf{9.6} & 318 & 68 & 62 & 389 & \textbf{59} \\
        \midrule
        \textbf{Mean} & 27.5 & 17.0 & \textbf{14.1} & 20.5 & 15.5 & 3.8 & 0.2 & \textbf{0.1} & 3.3 & 0.6 & 3.8 & 3.2 & \textbf{2.1} & 3.3 & 2.7 & 19.9 & 13.6 & \textbf{11.9} & 13.9 & 12.2 & 494 & 425 & \textbf{346} & 595 & 385 \\
        \bottomrule
    \end{tabular}
\end{table*}

\begin{figure*}[!t]
\centering
\includegraphics[width=\linewidth]{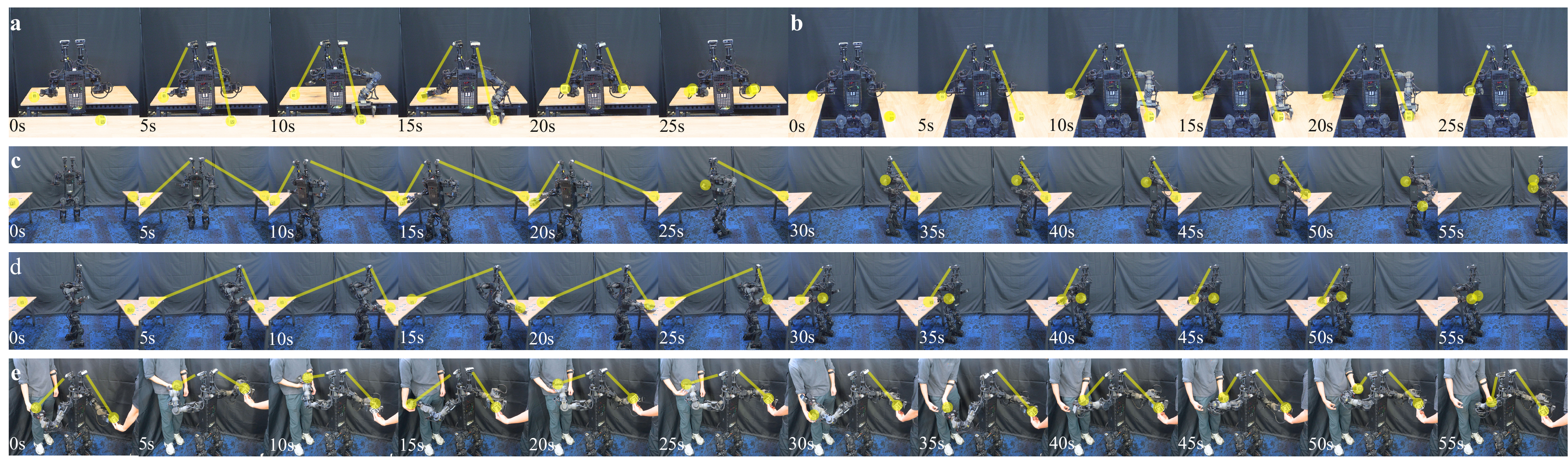}
\vspace{-20pt}
\caption{\textbf{Real-world reach-and-grasp.} Time-lapse sequences at 5~s intervals. \textbf{(a-b)}~The robot simultaneously observed and grasped targets in separated regions without reorienting its body: \textbf{(a)} front/back and \textbf{(b)} left/right. \textbf{(c-d)}~For far targets, the robot observed both targets, walked closer, and grasped the objects: \textbf{(c)} left/right far and \textbf{(d)} front/back far. Only the first grasp is shown in (c,d). Yellow circles mark the targets, and yellow lines show the two camera-to-target viewing directions. \textbf{(e)} The robot tracked and reached two moving objects held by two people. }
\label{fig:real-timelapse}
\end{figure*}

\subsection{Two-Target Reach-and-Grasp Benchmark}
\label{sec:results:task}
\label{sec:results:benchmark}
To evaluate whether VRW-guided camera configurations can improve visual manipulation, we tested actuated and fixed camera configurations (Act$_K$, Fix$_K$, $K=1,2$) along with Unitree G1 in six simulated two-target reach-and-grasp tasks. In each scenario, the robot needed to locate and grasp two objects (Fig.~\ref{fig:benchmark}), placed either to the left/right (L/R) or front/back (F/B) of the robot, with both targets on benches or one target on a bench and the other held by a human. For the bench scenes, the benches were either close to the robot ($0.20$~m), where the objects were within reach, or far from the robot ($0.8$~m), requiring the robot to walk before reaching for the targets.

The task was considered successful if the robot grasped both targets. \emph{Completion time} $\bar{T}$ was the total time to locate and retrieve both targets. The mechanical energy $E=\int\sum_j|\tau_j\dot q_j|dt$ was computed by integrating absolute joint power as an estimate of energy usage. $\bar{T}$ includes three parts. $\bar{T}_{\text{search}}$ is the time spent locating a target, through camera motion where possible, or by walking toward a known region where the target could be located. $\bar{T}_{\text{approach}}$ is the time spent walking toward a target outside the arm's reach. $\bar{T}_{\text{manip}}$ is the time spent reaching, grasping, and retracting along the planned arm trajectories.

We used the same heuristic planner to locate the target object across all camera configurations. The robot would either move its camera, rotate its body (fixed camera), or walk to the nearby table to discover the object. Once a target was within the camera field of view and arm reach, the robot began collision-aware grasp trajectory planning using cuRobo~\cite{curobo2023}. The arm trajectory planner ran asynchronously with the whole-body controller. We found that camera joint speeds above $2.0$~rad/s caused motion blur in the real world and made pose estimation unreliable, so we limited the maximum camera search speed to $2.0$~rad/s. The planned path was followed by a Jacobian tracker that compensated for gravity and robot body motion during reaching.
% Once the object was grasped, the robot retracted the arm to its default position.
The planner generated trajectories for simultaneous reaching if both targets were visible and reachable, or for single-target reaching if only one target was visible and reachable. All simulation trials were run in MuJoCo with full contact dynamics. The robots were controlled by our whole-body RL policy architecture (Section~\ref{sec:platform}).

Success rates were similar across the five configurations, ranging from $0.967$ to $0.994$. Compared with Fix$_2$, Act$_2$ reduced mean completion time by $17\%$ and energy by $19\%$ (Table~\ref{tab:scenarios-dynamic}). At $K=1$, Act$_1$ reduced completion time by $24\%$ and energy by $35\%$ relative to Fix$_1$.

Camera actuation reduced search time $\bar{T}_{\text{search}}$, approach time $\bar{T}_{\text{approach}}$, and manipulation time $\bar{T}_{\text{manip}}$ on average. Act$_2$ slightly reduced $\bar{T}_{\text{search}}$ relative to Fix$_2$, and Act$_1$ reduced it by $82\%$ relative to Fix$_1$ (Table~\ref{tab:scenarios-dynamic}). The reduction in search time mainly came from locating targets through camera motion rather than walking until the targets became visible. Camera actuation also reduced approach time because a far target could be located before walking, allowing the robot to walk directly toward the target rather than first toward the table. The second camera mainly increased pairwise coverage. Act$_1$ and Act$_2$ had similar single-target coverage $\tilde\eta$. Act$_2$ had higher pairwise coverage $\eta_2$ and could keep both targets in view for simultaneous reaching. In summary, camera configurations with higher visible-reachable and pairwise coverage reduced completion time and energy at similar success rates.

\mypara{Real-world deployment.} We deployed \ourrobot{} (Act$_2$) in all tabletop reach-and-grasp scenarios (i.e., front/back close, left/right close, left/right far, and front/back far) in the real world as shown in Fig.~\ref{fig:real-timelapse}. The two cameras observed the separated front/back or left/right targets, whereas the far scenarios additionally required the robot to walk closer before grasping. We also tested a dynamic scenario with two targets held and moved independently by two people (Fig.~\ref{fig:real-timelapse}\textbf{e}). During the trial, the two cameras tracked the separated moving targets while the robot followed them with its arms. When an arm occluded a target and tracking was temporarily lost, the corresponding camera actively searched for the target and reacquired it once it became visible. Our physical demonstrations show that larger VRW can support more effective visually guided manipulation while reducing the motion needed to acquire and maintain target views.
% https://docs.google.com/spreadsheets/d/1CqhkxLABD3MWVCth_Hf-cG08aGVekO0ez-8T5xh5xtA/edit?gid=0#gid=0

%==============================================================================
\section{Conclusion and Future Work}
\label{sec:conclusion}

Classical reachability does not capture whether a robot can perceive the workspace it can physically reach. We introduced the visible-reachable workspace (VRW) to couple these two capabilities, with a pairwise extension for concurrent observation of spatially separated regions. On \ourrobot, camera actuation increases visible-reachable coverage from $38\%$ to $97\%$, while a second independent view raises pairwise coverage from $0.45$ to $0.95$. In two-target manipulation, this improved sensing geometry reduces completion time and mechanical energy cost, with reductions in search, approach, and manipulation time. These results suggest that VRW is a useful design signal because improving visible-reachable coverage can translate into more efficient whole-body execution.

Future work can extend VRW beyond geometric visibility to incorporate sensing quality, uncertainty, and task-dependent perception requirements. We also plan to use the framework for automated co-design of sensing, morphology, and control across broader tasks and embodiments.

\bibliographystyle{IEEEtran}
\bibliography{references}

% Internal supplement. 
\ifinternalSupplement
\clearpage
\appendices
\section{RL details}
\label{app:internal-supplement}

\begin{samepage}
\subsection{Reward terms}
\begin{center}
    \footnotesize
    \tablecaption{Reward terms $r_i$ and weights $w_i$ for the reported direct controller ($r_t=\sum_i w_i r_i$). $a\!\to\!b$ = curriculum. $\langle\cdot\rangle$ = mean over joints/feet; $\ell,\ell_j$ = shared and joint-specific kernel widths; $\vect{v}_{xy},\omega_z$ = base linear/yaw velocity ($^{c}$ = commanded); $\vect{g}_{xy}$ = horizontal projected gravity; $\vect{q}^{0}$ = home pose, $\vect{q}^{\text{ref}}$ = moving reference; $p^z_f,\vect{v}^{xy}_f$ = foot height/planar velocity; $t^{\text{air}}$ = swing duration, $\Delta t^{\text{air}}{=}t^{\text{air}}_L-t^{\text{air}}_R$, $t_{\min}{=}0.05$~s, $t_{\max}{=}0.5$~s; $v_{\downarrow}{=}\max(0,-\dot p^z)$; $\mathbb{1}$ = indicator, $\mathbb{1}_{\text{land,valid}}$ = foot landing with two completed swings; $\bar q,\underline q$ = soft limits; $\tau_j,\bar\tau_j$ = joint torque/limit; $d_f$ = foot-flat gravity alignment; $h^\ast{=}0.1$~m, $v_0{=}0.3$~m/s, $\kappa{=}4$~s/m, $\sigma_\omega{=}0.5$~rad/s; $(x)_+{=}\max(x,0)$.}
    \label{tab:reward}
    \setlength{\tabcolsep}{5pt}
    \begin{tabularx}{\linewidth}{>{\raggedright\arraybackslash}p{2cm} >{\raggedright\arraybackslash}X >{\raggedleft\arraybackslash}p{2cm}}
        \toprule
        \textbf{Term} & \textbf{Expression $r_i$} & \textbf{$w_i$} \\
        \midrule
        \multicolumn{3}{l}{\emph{Task tracking}} \\
        Linear-velocity      & $\exp(-\lVert\vect{v}^{c}_{xy}-\vect{v}_{xy}\rVert^2/\ell^2)$ & $2.0$ \\
        Angular-velocity     & $\exp(-(\omega^{c}_z-\omega_z)^2/\ell^2)$ & $2.5$ \\
        Command-integral     & $\exp(-\lVert\vect{I}\rVert^2/\ell^2),\ \vect{I}{=}\!\int(\vect{c}{-}\vect{v})\,dt$ & $1.0$ \\
        Arm-pose             & $\exp(-\langle(\vect{q}_{\text{arm}}{-}\vect{q}^{\text{ref}}_{\text{arm}})^2\rangle/\ell^2)$ & $2.0$ \\
        Gaze                 & $\exp(-\langle(\vect{q}_{\text{cam}}{-}\vect{q}^{\text{ref}}_{\text{cam}})^2\rangle/\ell^2)$ & $1.0$ \\
        \midrule
        \multicolumn{3}{l}{\emph{Posture \& balance}} \\
        Upright              & $\exp(-\lVert\vect{g}_{xy}\rVert^2/\ell^2)$ & $1.0$ \\
        Variable posture     & $\exp(-\langle(\vect{q}{-}\vect{q}^{0})^2\rangle/\ell_j^2)$ & $1.0$ \\
        Foot flat orient.    & $\langle\exp(-(1-d_f^2)/\ell^2)\rangle_{\text{feet}}$ & $0.2$ \\
        \midrule
        \multicolumn{3}{l}{\emph{Gait shaping}} \\
        Feet air-time        & $\sum_f\mathbb{1}[t^{\text{air}}_f\!\in\![t_{\min},t_{\max}]]$ & $0.2$ \\
        Foot contact balance & $(\Delta t^{\text{air}})^2\,\mathbb{1}_{\text{land,valid}}\exp(-\lVert\vect{\omega}_{xy}\rVert/\sigma_\omega)$ & $-1.0$ \\
        Foot clearance       & $\sum_f|p^z_f{-}h^\ast|\tanh(\kappa\lVert\vect{v}^{xy}_f\rVert)$ & $-1.0$ \\
        Foot swing height    & $\sum_f(p^z_f{-}h^\ast)^2\,\mathbb{1}_{\text{land}}$ & $-0.25$ \\
        Foot slip            & $\sum_f\lVert\vect{v}^{xy}_f\rVert^2\,\mathbb{1}_{\text{contact}}$ & $-0.2$ \\
        Foot impact vel.     & $\sum_f\max(v_{\downarrow,f}^2{-}v_0^2,0)\,\mathbb{1}_{\text{contact}}$ & $-0.5$ \\
        \midrule
        \multicolumn{3}{l}{\emph{Regularization \& energy}} \\
        Action rate          & $\lVert\vect{a}_t-\vect{a}_{t-1}\rVert^2$ & $-0.5$ \\
        Arm action           & $\lVert\vect{a}^{\text{arm}}\rVert^2$ & $-0.01\!\to\!-0.5$ \\
        Base angular-vel.    & $\lVert\vect{\omega}^{\text{base}}_{xy}\rVert^2$ & $-0.05$ \\
        Joint-limit          & $\sum_j[\,(q_j{-}\bar q_j)_+ + (\underline q_j{-}q_j)_+\,]$ & $-1.0$ \\
        Self-collision       & $N_{\text{self-contact}}$ & $-1.0$ \\
        Joint torque         & $\sum_j(|\tau_j|-\tfrac12\bar\tau_j)_+^2$ & $-2{\times}10^{-3}$ \\
        Joint power          & $\sum_j|\dot q_j\,\tau_j|$ & $-4{\times}10^{-4}$ \\
        \bottomrule
    \end{tabularx}
\end{center}
\end{samepage}

\begin{samepage}
\subsection{Observation space}
\begin{center}
    \footnotesize
    \tablecaption{Observation space. The actor stacks the listed noisy, delayed $120$-D frame over $20$ steps; the asymmetric critic receives the current clean $169$-D observation. The top block is shared; the bottom block is critic-only. Noise is additive per step ($\mathcal{U}$ uniform, $\mathcal{N}$ Gaussian); joint position also has a per-episode $\mathcal{U}(\pm0.05)$ bias, and both joint position and joint velocity have a $0$--$2$-step random delay.}
    \label{tab:obs}
    \begin{tabularx}{\linewidth}{>{\raggedright\arraybackslash}X c >{\raggedright\arraybackslash}p{3.1cm}}
        \toprule
        \textbf{Term} & \textbf{Dim} & \textbf{Noise / delay} \\
        \midrule
        Base angular velocity (IMU)   & 3  & $\mathcal{N}(0,0.1)$ \\
        Projected gravity (IMU)       & 3  & $\mathcal{U}(\pm0.05)$ \\
        Joint position                & 31 & $\mathcal{U}(\pm0.01)$, bias, delay \\
        Joint velocity                & 31 & $\mathcal{U}(\pm1.0)$, delay \\
        Last action                   & 31 & --- \\
        Command (twist $\vect{c}_t$)  & 3  & --- \\
        Arm reference $\vect{q}^{\text{ref}}_{\text{arm}}$ & 14 & --- \\
        Gaze reference $(\psi^\ast,\theta^\ast)_{L,R}$ & 4 & --- \\
        \midrule
        \multicolumn{3}{l}{\emph{Privileged (critic only)}} \\
        Command integral $\int(\vect{c}_t-\vect{v}_t)\,dt$ & 3 & --- \\
        Base linear velocity          & 3  & --- \\
        Foot height / air-time / contact & 2/2/2 & --- \\
        Foot contact forces           & 6  & --- \\
        Joint torque                  & 31 & --- \\
        \bottomrule
    \end{tabularx}
\end{center}
\end{samepage}

\begin{samepage}
\subsection{Training hyperparameters}
\begin{center}
    \footnotesize
    \tablecaption{FlashSAC training hyperparameters. Off-policy SAC with a distributional (C51) asymmetric critic and RMA-CNN actor-history encoder.}
    \label{tab:hyperparams}
    \begin{tabularx}{\linewidth}{>{\raggedright\arraybackslash}X r @{\hspace{1.5em}} >{\raggedright\arraybackslash}X r}
        \toprule
        \textbf{Parameter} & \textbf{Value} & \textbf{Parameter} & \textbf{Value} \\
        \midrule
        Parallel envs        & $4096$   & Discount $\gamma$          & $0.99$ \\
        Steps/env per iter   & $24$     & Target smoothing $\tau_{\text{target}}$    & $0.01$ \\
        Iterations           & $15\,000$ & Actor / critic LR         & $3\!\times\!10^{-4}$ \\
        Replay buffer/env    & $256$    & Temperature LR $\alpha$    & $2\!\times\!10^{-4}$ \\
        Batch size           & $4096$   & $\alpha_{\text{init}}$ (auto-tuned) & $0.01$ \\
        History length       & $20$     & Target entropy $\mathcal{H}_{\mathrm{target}}$    & $0.15$ \\
        Encoder embed / latent & $32/128$ & Critic nets (C51 atoms)  & $2\times101$ \\
        MLP trunk            & $512/256/128$ & Critic support $[v_{\min},v_{\max}]$ & $[-5,5]$ \\
        \midrule
        \multicolumn{4}{l}{\emph{Actor / critic observation groups}} \\
        Actor observation (deployed) & $120$ & Critic observation (privileged) & $169$ \\
        \bottomrule
    \end{tabularx}
\end{center}
\end{samepage}

\begin{samepage}
\subsection{Domain randomization}
\begin{center}
    \footnotesize
    \tablecaption{Domain randomization. Applied at episode reset (R) or at fixed time intervals (I). Scale = multiply nominal; add = additive offset.}
    \label{tab:dr}
    \begin{tabularx}{\linewidth}{>{\raggedright\arraybackslash}X >{\raggedright\arraybackslash}p{3.3cm} c}
        \toprule
        \textbf{Parameter} & \textbf{Range} & \textbf{When} \\
        \midrule
        Foot friction        & $[0.3,2.5]$ (abs) & I \\
        Foot contact impedance (\texttt{solimp}) & zero-penetration $d_0\,[0.85,0.99]$, transition width $[2,10]$~mm & I \\
        Foot contact response (\texttt{solref}) & time constant $\tau_{\text{contact}}\,[15,25]$~ms, damping ratio $\zeta\,[0.9,1.1]$ & I \\
        PD gains $k_p,k_d$   & $\times[0.9,1.05]$ & R \\
        Body mass            & $\times[0.85,1.15]$ & I \\
        Base CoM offset      & $\pm1$~cm (add) & I \\
        Wrist EE CoM offset  & $\pm5$~mm (add) & I \\
        Joint friction loss  & $\pm0.01$ (add) & R \\
        Joint damping        & $[0,0.01]$ (add) & R \\
        Push (lin./ang.\ vel.) & $\pm0.5$~m/s, $\pm0.5$~rad/s & I ($5$--$15$~s) \\
        Init.\ pose / joints & yaw $[-\pi,\pi]$, $\pm0.1$~rad & R \\
        \bottomrule
    \end{tabularx}
\end{center}
\end{samepage}

\subsection{Command sampling}
Yaw is driven in heading mode with stiffness $0.5$, and heading is sampled over $[-\pi,\pi]$. In 20\% of environments, the robot is commanded to stand. Commands are resampled every $2$--$8$~s.

\fi
\end{document}